\documentclass[man,floatsintext]{apa7}

\usepackage{pslatex}
\usepackage{apacite}
\usepackage{graphicx}
\usepackage[rightcaption]{sidecap}
\usepackage{natbib}

\title{Assessing the alignment between infants' visual and linguistic experience\\using multimodal language models}
\shorttitle{Infants' Vision--Language Alignment}

\authorsnames[1,2,2,1,1,3,1,1,2]{Alvin W.M. Tan*, Jane Yang*, Tarun Sepuri, Khai Loong Aw, Robert Z. Sparks, Zi Yin, Virginia A. Marchman, Michael C. Frank, Bria Long}
\authorsaffiliations{{Department of Psychology, Stanford University},
{Department of Psychology, University of California, San Diego},
{Department of Psychology, Tsinghua University}}

\abstract{Figuring out which objects or concepts words refer to is a central language learning challenge for young children. Most models of this process posit that children learn early object labels from co-occurrences of words and their referents that occur when someone around them talks about an object in the immediate physical environment. But how aligned in time are children’s visual and linguistic experiences during everyday learning? To date, answers to this question have been limited by the need for labor-intensive manual annotations of vision--language co-occurrences. Here, we evaluate the use of contrastive language-image pretraining (CLIP) models to automatically characterize vision--language alignment in egocentric videos taken from the infant perspective in home environments. After validating CLIP alignment scores using human alignment judgments, we apply this metric to a large corpus of infant-perspective videos. We show that idealized aligned moments for learning (e.g., ``look at the \textit{ball}'' with a ball present in the child's view) are relatively rare in children's everyday experiences compared to modern machine learning datasets, and highlight variability in alignment both within and across children. These findings suggest that infrequent alignment is a constraint for models describing early word learning and offer a new method for investigating children's multimodal environment. 
}

\keywords{early word learning; head-mounted cameras; multimodal language models; naturalistic observations}

\authornote{Correspondence at \texttt{tanawm@stanford.edu}}

\begin{document}

\maketitle

Learning that the word ``dog'' refers to golden retrievers and Chihuahuas, but not cows, requires children to form visual concepts from their everyday experiences. Children's learning environments provide the basis for these inferences about which visual concepts words refer to, but how often do children actually see clearly labeled exemplars of categories, and how important are these moments for forming visual concepts? Despite the centrality of this question for theories of early word learning, we have limited data about how temporally aligned children's visual and linguistic experiences are during development. Here, we leverage contrastive language--image pre-training (CLIP) models \citep{radford2021learning} to assess how temporally aligned children's visual and linguistic experiences are over early development. 

Associative models of early word-learning---such as cross-situational word learning---predict that children extract meaning from co-occurrences of frequent words and objects in their environment \citep{yu2007rapid, zhang2021cross}. Egocentric videos taken from the child's perspective offer an unprecedented window to children's learning environments during their everyday experiences \citep{yoshida2008s}, allowing researchers to quantify how often children actually experience word--object co-occurrences. Initial investigations suggest that the more that children experience a referent (as measured by prevalence in egocentric videos), the more likely they are to learn the word for it in a subsequent looking-while-listening task \citep{bergelson2017nature}. Recent computational simulations have further highlighted the importance of clearly labeled examples in building useful category representations: \citet{vong2024grounded} found that CLIP models can learn rudimentary representations from roughly 60 hours of a single child's naturalistic head-mounted data, but that directly labeled referents were disproportionately more informative for learning. 

Understanding the degree to which children's visual and linguistic experiences are aligned in everyday learning is thus critical for both building realistic models of early word learning and understanding variability in learning rates across children. Early language learning is highly variable between children \citep{frank2021}, and this variability may have downstream consequences for both language and reading success \citep{marchman2008}. To date, however, our knowledge has been limited by the need for manual annotations \citep{bergelson2017nature, yoshida2008s, frank2013}, which are unfeasible beyond a few hours of data. Here, we overcome this barrier to progress by using multimodal language models to examine infants’ visual and linguistic experience in a large, naturalistic dataset of egocentric visual experience \citep{long2024babyview}. 

\begin{figure}
\centering
\includegraphics[width=\textwidth]{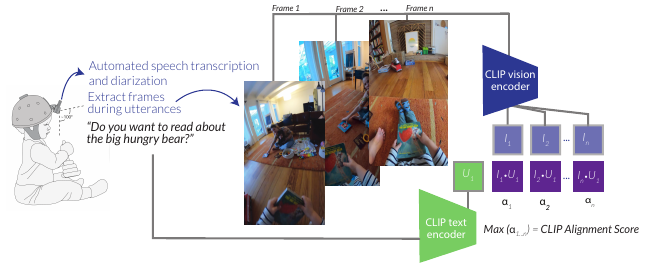}
\caption{Overview of the pipeline for CLIP alignment scores, showing example frames from a BabyView camera in a naturalistic home environment; CLIP alignment scores are calculated by taking the dot product of the normed embeddings from the CLIP text and vision encoders for each frame associated with an utterance (as transcribed by Distil-Whisper). We then take the maximum alignment score (visualized in purple) for each utterance.} 
\label{fig:methods-figure}
\end{figure}

Specifically, we leverage advances in multimodal large language models---contrastive language--image pre-training models \citep{radford2021learning}---to quantify the degree to which what children see is temporally aligned with the semantic content in the language that they hear. These multimodal models contain jointly trained visual and language encoders that maximize similarity between images and their captions in large datasets. For example, the language encoder could represent an utterance (i.e., ``Do you want to read about the big hungry bear?'') and the vision encoder could represent a concurrent image (i.e., a living room scene with a book in the foreground, see Figure~\ref{fig:methods-figure}); computing cosine similarity of the resulting embeddings from both encoders provides a metric of semantic alignment between visual and linguistic inputs.

However, to date no work has systematically examined whether this metric can capture human judgments about the degree to which utterances and images are aligned. We address this gap by first validating CLIP alignment scores against human performance in a 4-alternative forced choice (4AFC) matching task using a stratified sample of utterances and frames from BabyView, a new open dataset of egocentric videos \citep{long2024babyview} with automatic transcriptions and speaker diarization. After establishing that CLIP alignment scores correspond to human judgments, we apply this same metric to concurrent frames that occur during each utterance in the entire dataset, calculating an alignment metric for every utterance. 

With these alignment estimates, we quantify consistency and variability in how children’s visual and linguistic experiences are aligned across contexts and individuals. Overall, we find that the proportion of children's everyday experiences where the visual and linguistic inputs are highly aligned is relatively low compared to modern machine learning datasets. We then examine variation in vision--language alignment across developmental age, within individual children, across activity contexts (e.g., reading, eating), and depending on the linguistic content of the utterances. By integrating computational innovations to describe development at scale, we aim to contribute to a broader and more inclusive view of everyday learning contexts, with the ultimate goal of understanding how variability across learning environments influences learning outcomes in this population.

\section{Dataset \& Methods}

We analyze naturalistic, egocentric video from children's perspective, collected with the BabyView camera \citep{long2024babyview}. Compared to prior cameras \citep{sullivan2021saycam}, the BabyView camera offers substantially better audio quality (increasing the feasibility of automatic transcriptions), and a wider vertical field of view (increasing the likelihood of referents being visible). We analyze BabyView dataset releases 2025.1 and 2025.2, including only monolingual English-speaking families, yielding 325 hours of speech from the families of 19 children (5--36 months of age).

\subsection{Automated transcriptions}
All videos were transcribed using Distil-Whisper \citep{radford2022robust, gandhi2023distil}, specifically the \texttt{distil-large-v3} model,\footnote{\url{https://huggingface.co/distil-whisper}} with a word error rate of .35 (see \citet{long2024babyview} for further details on the transcripts). We conducted analyses both on all utterances in the videos as well as only those utterances that were identified as being produced by an adult by a speaker diarization model \citep{lavechin2020open}. 

\subsection{CLIP alignment calculation}
To calculate vision--language alignment scores, we used an implementation of the OpenAI CLIP ViT/B32 model \citep{radford2021learning, jina2023clip}.\footnote{\url{https://github.com/jina-ai/clip-as-service}} For each utterance, we calculated the cosine similarity between the image embedding for each frame (sampled at 1 fps) with the text embeddings for each utterance (``You want the ball''); we refer to this as the \textit{CLIP alignment score}. For each utterance, the CLIP alignment score was the maximum score that occurred in any frame occurring during that utterance. See Figure~\ref{fig:methods-figure} for an overview of this calculation.

\subsection{Activity \& location annotations}
We additionally annotated the frames with activity and location annotations from \cite{sepuri2025characterizing}. These annotations were obtained by chunking each video into 10-second clips, and prompting a video question-answering model \citep[VideoLLaMA 3,][]{zhang2025videollama} to select the appropriate activity and location for each clip from a set of options. These annotations were used to sample a subset of frames for the human validation task described below, ensuring broad coverage of the dataset.

\subsection{Human annotations of alignment}
We validated CLIP alignment scores through a four-alternative forced choice (4AFC) human annotation task.\footnote{Preliminary efforts directly annotating image--utterance pairs for alignment (on a 1--5 Likert scale) suggested that this method was likely to be noisy due to the imprecision of what ``alignment'' means.} We developed two versions of a forced-choice task that was similar to the contrastive training objective of CLIP models, which evaluated the alignment of the visual and speech information. In one condition, annotators were presented with an utterance and asked to guess which of four frames matched the utterance (\textit{image} condition). In a second condition, annotators were presented with one frame and asked to guess which of four utterances matched the frame (\textit{utterance} condition).

To construct the set of trials shown to annotators, we conducted two types of stratified sampling across utterances. First, we stratified across CLIP alignment scores---we binned each utterance score into one of 5 bins and sampled up to 80 frames in each bin. Next, we stratified across the detected activities and locations \citep{sepuri2025characterizing}, sampling at least 10 frames for each activity and for each location separately. Our goal in using these location and activity locations was to ensure as broad of coverage as possible over this very large dataset while only selecting a small set of utterances and their corresponding frames for analyses. Together, there were 732 frame--utterance pairs sampled for annotation. The three distractors for each trial were randomly sampled from the remaining frames and utterances within the set of sampled pairs. 

We recruited 80 English-speaking annotators who completed a web-based version of this task on Prolific. Each annotator saw a mean of 110 test trials in one of the two conditions ($\sim$6 annotations per condition per trial), as well as 5 catch trials containing simple vocabulary questions. Annotators were excluded if they answered more than one catch trial incorrectly ($N = 2$). All annotators were blind to the purposes of the study and reported that they were English-speaking adults residing in the United States. 

\section{Model Validation}

\subsection{Validation via human annotations}
We first assessed the feasibility of using the alignment scores by manually examining a random selection of frames with high alignment scores (CLIP score $\geq$ .24, used in prior work \citep{vong2024grounded}). We found qualitative evidence that utterances with high alignment tended to refer to a concrete object that was visible (e.g., ``Don't bite your toys'', ``There's your green chair'', ``Cornflakes''). In particular, several high alignment frames came from books visible in the frame that caregivers were reading aloud; thus, there was text in the frame that exactly matched caregivers' utterances. Some mismatches occurred when there was a contextual match (e.g., in a kitchen scene, the \textit{fork} was absent for the utterance ``Should I get you a fork?''), such that the alignment score was relatively high despite the referent being absent. However, manual inspection suggested that many utterances that had low alignment tended to refer to absent entities (``when we go back to Ohio, you'll see lots of cows.'') or were directed towards other adults (``…down by the station later in the evening.'')

\begin{figure}
    \centering
    \includegraphics[width=\textwidth]{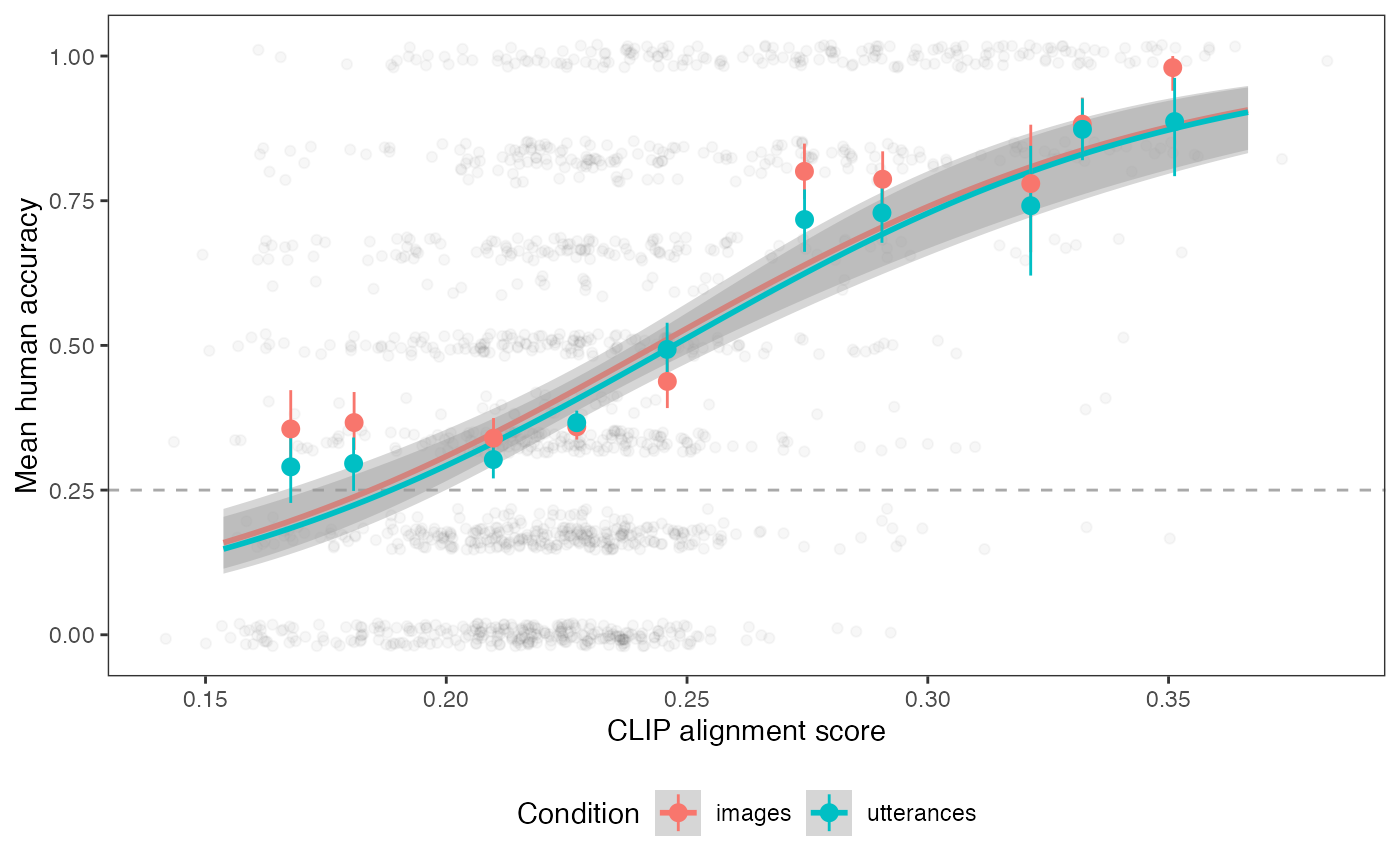}
    \caption{Human 4AFC accuracy by condition as a function of CLIP alignment score. Error bars indicate bootstrapped 95\% confidence intervals by CLIP score decile, and lines indicate best-fit logistic curves along with their 95\% confidence band. Dashed line indicates chance-level performance.}
    \label{fig:validation}
\end{figure}

Next, we systematically evaluated the correspondence between human and model judgments by examining the human performance for frames on our 4AFC matching task; here, ground truth was the original frame--utterance pairing. We examined how well annotators could match each frame to an utterance (and vice versa) as a function of the alignment score produced by CLIP. As shown in Figure~\ref{fig:validation}, we found that human performance scaled with the alignment score, such that frame--utterance pairs with higher CLIP scores were also more likely to be correctly guessed by humans. This relationship was confirmed by a logistic regression predicting human accuracy as a function of CLIP score, condition, and their interaction. CLIP score emerged as a significant predictor of human accuracy ($b = 18.5$ [14.0, 23.1], $p < .001$). Neither condition nor the interaction between CLIP score and condition were significant predictors ($p > .8$ for both). This result suggests that the image and utterance conditions returned very similar values. At the utterance level, human accuracies showed a significant correlation between the two conditions ($r = 0.541$, $p < .001$). The predicted intercept (accuracy = 0.5) of the logistic model occurred at a CLIP score of 0.25; this value was close to the threshold of 0.24 for high alignment used in previous work \citep{vong2025robustness}.

\begin{figure}
    \centering
    \includegraphics[width=\textwidth]{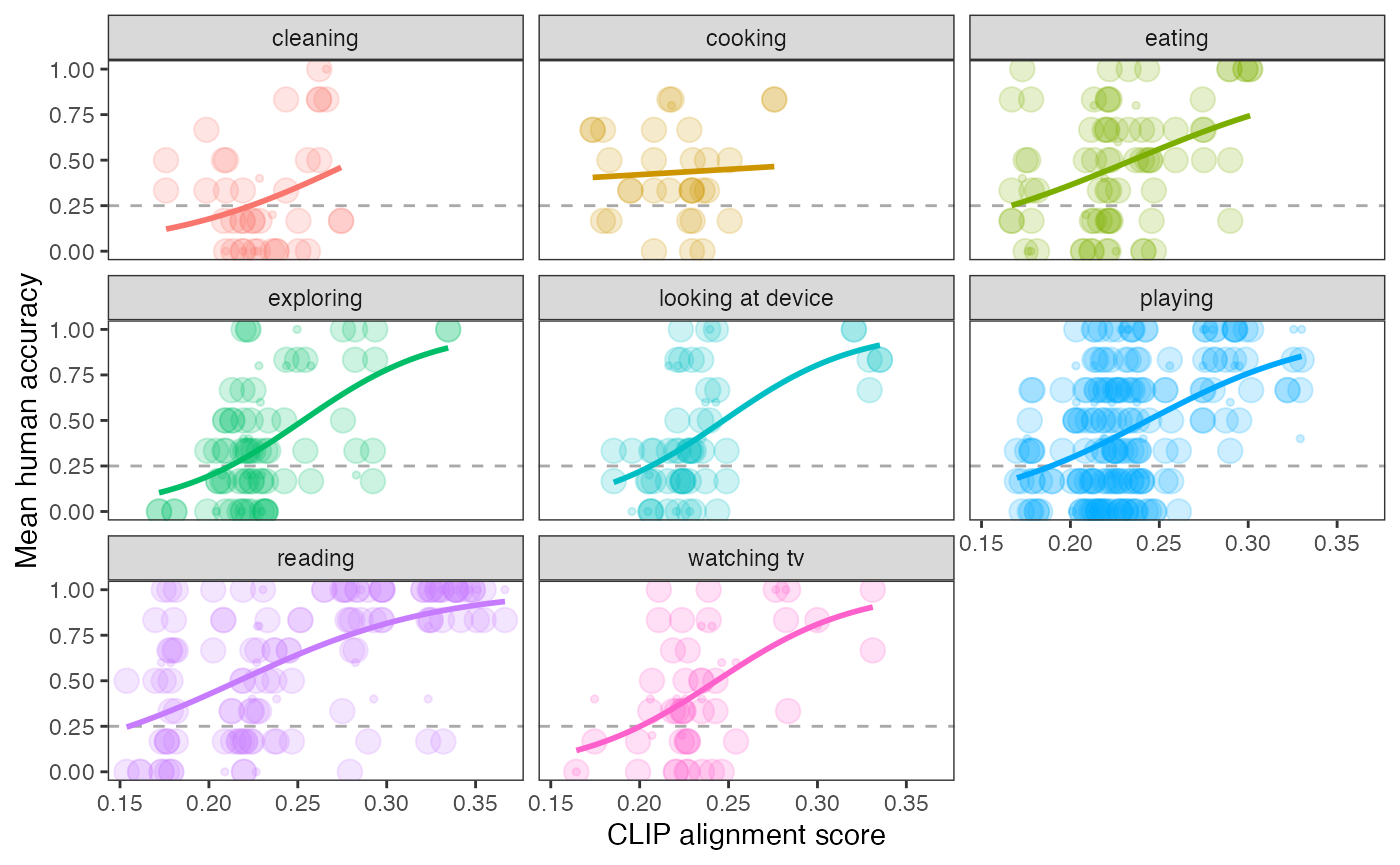}
    \caption{Human 4AFC accuracy as a function of CLIP alignment score, for the 8 most frequent activities. Lines indicate best-fit logistic curves. Dashed line indicates chance-level performance.}
    \label{fig:activity}
\end{figure}

The overall relationship between CLIP alignment score and accuracy also held across different activities, as shown in Figure~\ref{fig:activity}. While there were some observed differences in the intercept of the different logistic curves, these points occurred within the interval (0.24, 0.26).

\begin{figure}
    \centering
    \includegraphics[width=\textwidth]{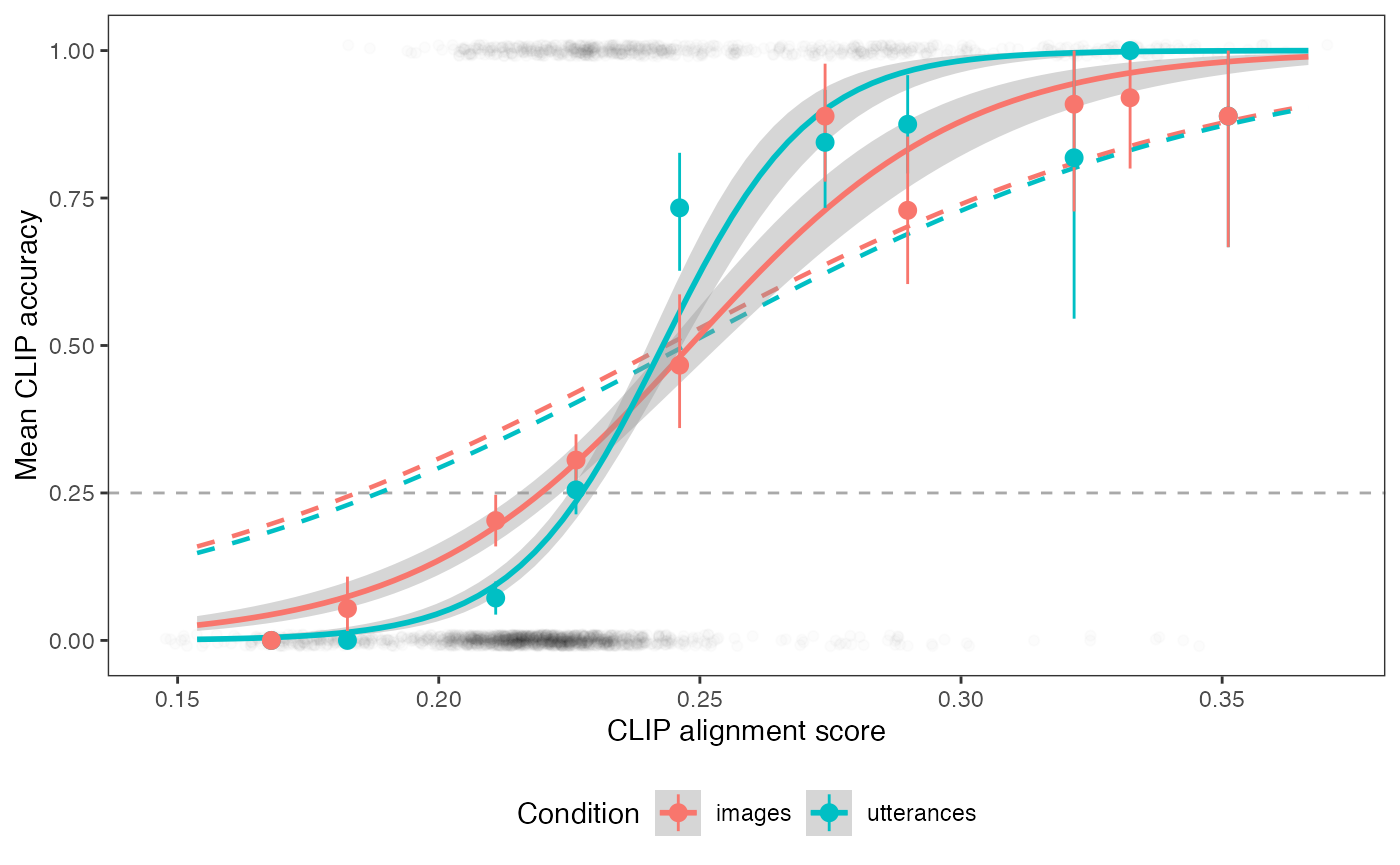}
    \caption{CLIP 4AFC accuracy as a function of CLIP alignment score. Erorr bars indicate bootstrapped 95\% confidence intervals by CLIP score decile, and lines indicate best-fit logistic curves along with their 95\% confidence band. Dashed colored lines reflect human 4AFC accuracies (as in Figure~\ref{fig:validation}). Dashed grey line indicates chance-level performance.}
    \label{fig:clip}
\end{figure}

\subsection{Interpreting model alignment scores vs. model accuracy}

We explored a set of analyses to help interpret CLIP alignment scores relative to human CLIP accuracy scores. We evaluated the same CLIP model (CLIP ViT/B32) on the same 4AFC task presented to our human annotators. 
We operationalized model choice as the softmax over the alignment scores between an image and the four possible utterances (or vice versa). Accuracy was then determined by whether the model chose the correct target. 
We observed a positive relationship between alignment scores and 4AFC accuracy. This relationship also provides an intuitive interpretation of CLIP score: for a given image–-utterance pair with a particular CLIP score and three distractor pairs, it approximates the probability that the CLIP model will select the correct pair.


Additionally, we found an interesting difference between CLIP and humans with respect to the shape of the relationship between CLIP alignment scores and accuracy. At low CLIP alignment scores, the CLIP model is almost at 0\% accuracy, followed by a steep increase in accuracy at medium CLIP scores, and close to 100\% accuracy at high CLIP scores (see Figure~\ref{fig:validation}). In contrast, humans show approximately chance-level performance at low CLIP scores, and accuracy increases with a shallower slope. Further, humans show above-chance performance at a lower threshold than the CLIP model. This difference suggests that CLIP may be wrongly assigning low cosine similarity to frame--utterance pairs on which humans succeeded in the 4AFC task. Nonetheless, the intercepts for both humans and models lie within the interval (0.24, 0.26), further validating these values as a threshold for high alignment.

\section{Descriptive Results \& Analyses}

Next, we used per-utterance CLIP alignment scores to quantify the consistency and variability in visual--linguistic alignment in a large corpus of egocentric videos taken from the infant perspective. Below, we analyze these alignment scores across the overall dataset, within individual children's data, and with respect to the content of the linguistic utterance.

\begin{figure}
    \centering
    \includegraphics[width=\textwidth]{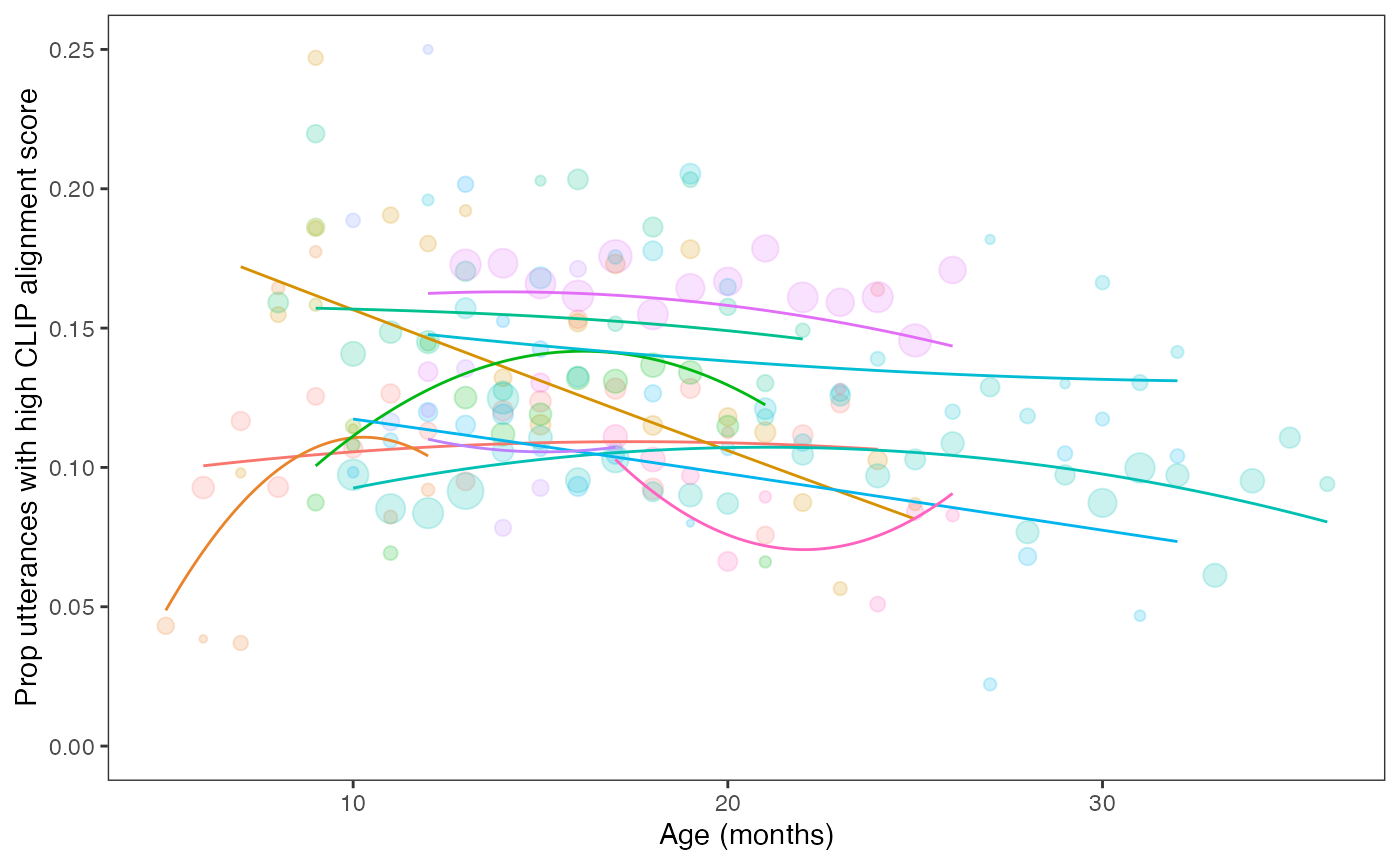}
    \caption{Proportion of highly-aligned adult utterances as a function of the child's age. Colors reflect individual children, and lines reflect best-fit LOESS curves; the size of the dots scales with the amount of utterances in each age bin. }
    \label{fig:age}
\end{figure}

\subsection{Vision--language alignment across individuals}

Overall, we found that the children's everyday experiences contained relatively infrequent moments with high visual--linguistic alignment, with considerable variability across individuals. Figure~\ref{fig:age} shows the proportion of adult utterances in a given video that had high alignment, as a function of the age of the child at the time of recording; we found the same trend whether we used raw cosine similarity scores or these thresholded scores, as per \citet{vong2025robustness}. Notably, ``highly aligned'' frame--utterance pairs occurred less than 20\% of the time for most children at almost all ages, and the overall average across our entire dataset was $M$ = 12.64\%. However, note that there was also considerable heterogeneity across families, with some families trending around 16\% high alignment while others were closer to 8\%, a two-fold difference.  

\begin{figure}
    \centering
    \includegraphics[width=\textwidth]{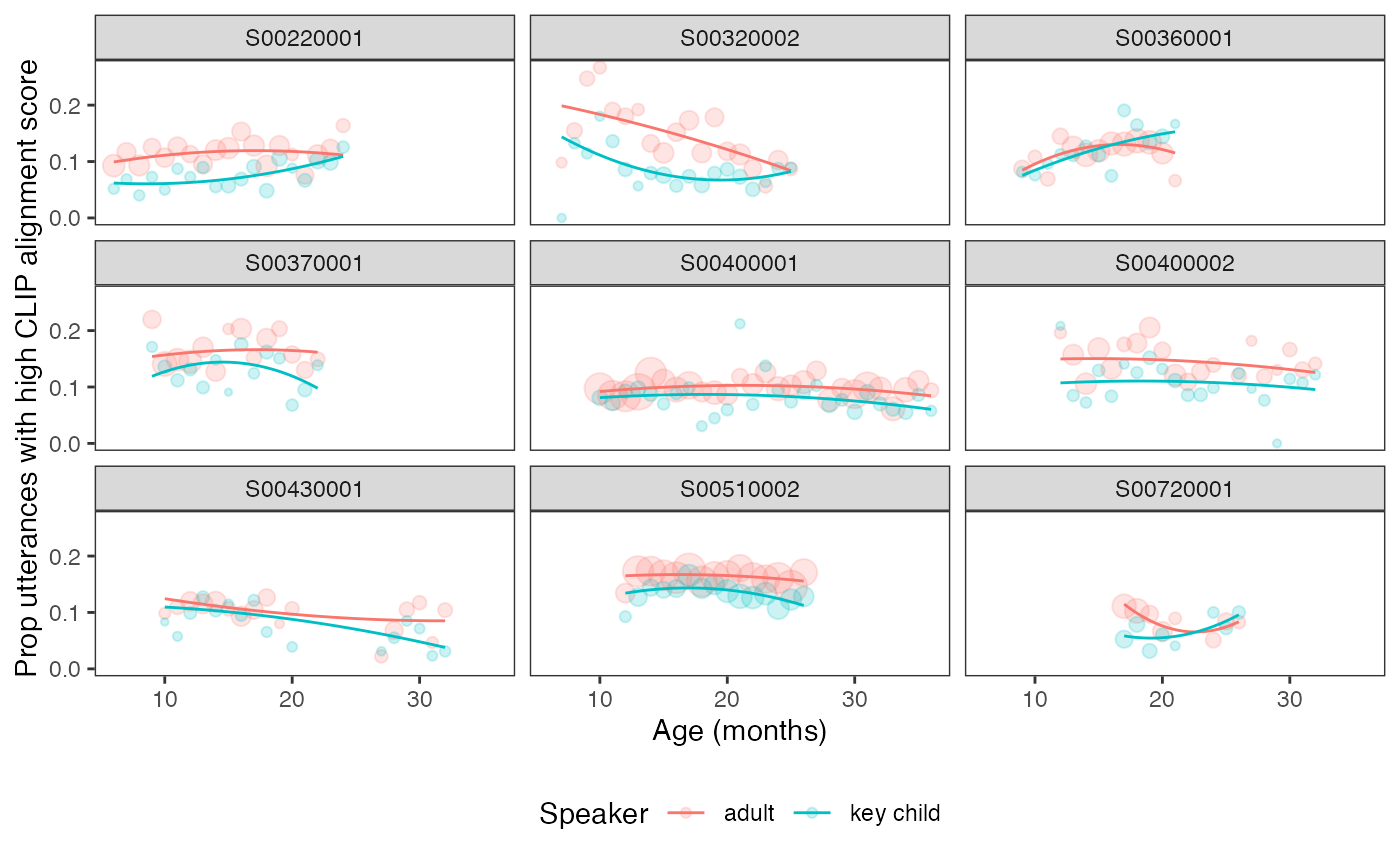}
    \caption{Proportion of highly aligned utterances as a function of speaker and the child's age, for the 9 families with the most data. Lines represent best-fit LOESS curves.}
    \label{fig:speaker}
\end{figure}

\subsection{Higher alignment in adult-produced speech}

What drives higher CLIP alignment scores?  We examined several possible sources of variance in the alignment scores that we observed. First, we anticipated that utterances produced by adults would have higher alignment than utterances produced by either the child who was recording (i.e., ``key child'') or other children in the household. We suspected this would be the case given both higher automated transcription accuracies for adult-produced speech \citep{long2024babyview} and the general tendency of caregivers to perhaps try to label or describe a child's visual environment for them.\footnote{The current state-of-the-art diarization algorithm that we employed \citep{lavechin2020open} is unable to classify whether adult speech is child-directed; however, it is able to distinguish between child-produced and adult-produced speech.} Thus, we next examined whether we saw systematic differences in alignment for speech that contained adult- and key child-produced speech, as shown in Figure~\ref{fig:speaker}. As expected, we found that adult-produced speech had higher CLIP alignment scores, on average. This finding was supported by a linear mixed effects model predicting the proportion of highly aligned utterances as a function of age, speaker, and their interaction, along with random intercepts and age slopes by child. In this model, speaker type was the only significant predictor ($b = 0.039$ [0.018, 0.060], $p < .001$). Thus, the higher alignment in adults vs. children was consistent regardless of the age of the child at the time of recording. 

\begin{figure}
    \centering
\includegraphics[width=\textwidth]{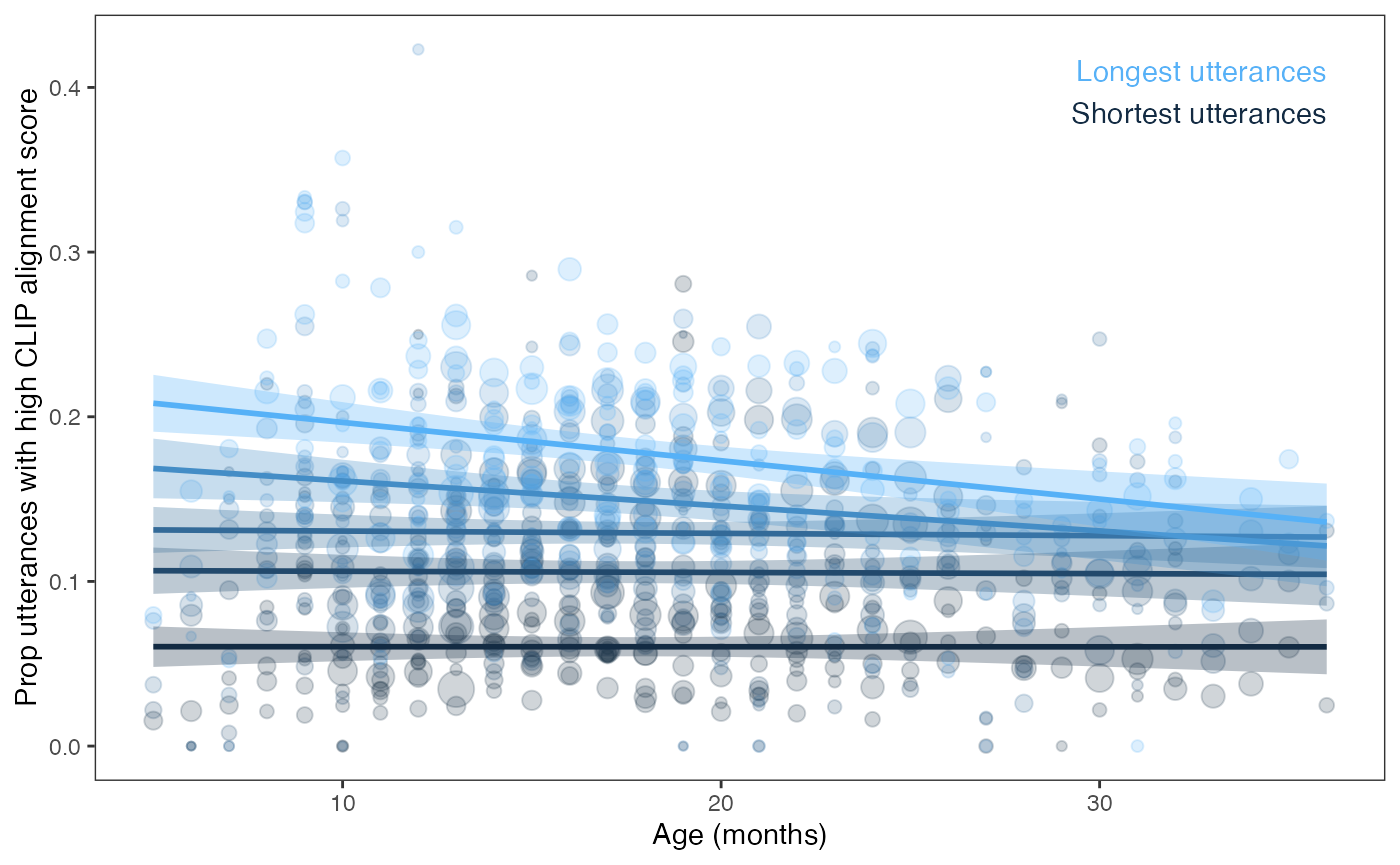}
    \caption{Proportion of utterances with high vision--language alignment scores as a function of children's age during recording and utterance duration; here, we restrict our analyses to adult-produced speech. Utterance duration is binned into quintiles; lines reflect best-fit linear trend lines for each utterance length decile; dot size reflects the number of utterances in each age/utterance duration bin. Lighter lines/dots refer to  utterances of longer duration.}
    \label{fig:length}
\end{figure}

\subsection{Higher alignment for longer utterances}

We next examined the possible role of utterance duration in driving alignment scores, as shown in Figure~\ref{fig:length}. Analyzing adult utterances only, we found that utterances that spanned a longer duration had higher CLIP alignment scores than shorter utterances, on average. This relationship was confirmed using a linear mixed effects model predicting CLIP alignment score as a function of utterance length, age, and their interaction, along with random slopes for utterance length and intercepts by child (random slopes for age failed to converge). In this model, there was a significant effect of utterance duration ($b = 9.460 \times 10^{-4}$ [$7.920 \times 10^{-4}$, $1.100 \times 10^{-3}$], $p < .001$). That is, longer utterances (e.g., ``Here are your blue shorts with the dinosaurs'') tended to have higher vision--language alignment than shorter utterances (e.g., ``Hi baby'', ``Your bottle''). In addition, vision--language alignment decreased slightly across age for \textit{longer} but not shorter utterances, which had overall lower alignment, as evidenced by an interaction in this same model between utterance duration and age ($b = -1.072 \times 10^{-5}$ [$-6.567 \times 10^{-6}$, $-1.487 \times 10^{-5}$], $p < .001$). Thus, these exploratory analyses suggest that adult speech may decrease slightly in its overall visual--linguistic alignment across age as children become older---perhaps reflecting changes in care routines or in the degree to which adults are referencing absent or future objects and events.

\subsection{Higher alignment for utterances containing more frequent and concrete lemmas}

We also investigated the possible role of the content of the utterances in driving alignment, focusing on properties of individual words in the utterances. We estimated the CLIP alignment score per word by lemmatizing all utterances using UDPipe \citep{straka2016udpipe}. Then, for each lemma, we calculated the mean CLIP alignment score for all utterances containing that lemma. We then filtered down to lemmas that appeared in at least 10 different utterances to remove idiosyncratic low-frequency items, retaining 1213 lemmas ($\sim$20\%). We then merged these lemmas with several psycholinguistic predictors: \textit{log frequency} of occurrence of the lemmas in the BabyView dataset, \textit{concreteness} \citep{brysbaert2014concreteness}, \textit{imageability} \citep{cortese2004imageability, schock2011imageability}, and \textit{sensorimotor} and \textit{action strength} \citep{lynott2020lancaster}. These psycholinguistic predictors were chosen as they may relate to the extent to which the meaning of words can be represented with elements in the visual field, and that have been shown to affect word processing in adults \citep{khanna2021how}. 


Because the psycholinguistic norms did not cover all the lemmas present in our dataset, we used multiple imputation (with 5 imputations) using predictive mean matching to handle missing data. We then fit a linear model predicting lemma mean CLIP alignment score, with the fixed effects of log frequency, concreteness, imageability, sensorimotor strength, and action strength. Only two predictors emerged as significant. More concrete lemmas tended to occur in utterances with higher CLIP alignment scores than less concrete lemmas ($b = 0.0016$ [0.0007, 0.0025], $p < .001$), and lemmas that were more frequent occurred in utterances with higher CLIP alignment scores than less frequent lemmas ($b = 0.0007$ [0.0003, 0.0010], $p < .001$). 
We do note, however, that lemmas seem to follow a Zipfian distribution in frequency, thus there are a few highly frequent lemmas that may be driving most of the effect of log frequency.

\section{Discussion}

Can machine learning models be used to explore the alignment of linguistic and visual experience in children's everyday experiences? To answer this question, we first validated a computational technique for assessing alignment in videos of children's experience. We found that the similarity of the linguistic and visual embeddings in a contrastive language--image pre-training model (CLIP) was related to empirical human judgments of vision--language alignment, suggesting that this metric can be used as a proxy for vision--language alignment in other datasets.

Using this validated approach, we then asked: How aligned in time is what children see with what they hear? While idealized moments of joint attention between children and their caregivers may produce some highly aligned moments, we found that such aligned moments occurred with considerable variability across individuals and time---at most every one in five utterances in any given age bin, and far less often in many cases. Instead, children often heard speech between two caregivers talking with each other, from another child, or referring to an object or situation that was not in the ``here and now'' (e.g., ``Let's go to the park''). Notably, this lower level of alignment contrasts with the amount of visual--linguistic alignment in curated datasets used to train vision--language models (e.g., COCO \citep{lin2015microsoft} or Flickr \citep{young2014image}). Furthermore, the considerable heterogeneity across families in alignment rates highlights that young children's word learning mechanisms need to be relatively robust to fairly large differences in the alignment of the visual and linguistic information.

A few different features significantly predicted CLIP alignment scores. First, adult-produced speech was significantly more aligned than child-produced speech. There are several possible reasons for this finding---for example, adult speech may have been more contextually contingent, or the adults may have been engaging in more pedagogical talk that was directly tied to the immediate context. It is also possible that adult utterances were simply more well-structured or better-transcribed compared to child utterances, and thus, more similar to the training data of CLIP. A related finding is that longer utterances exhibited higher alignment---and adults, given their better command of language, typically produce longer utterances than children. Future work could potentially disentangle these hypotheses by manipulating adult and child utterances in various ways to determine the impact of register, sentence structure, and context contingency on alignment.

Additionally, some types of lemmas occurred more often in highly aligned utterances than other lemmas. In particular, more frequent lemmas occur in more aligned utterances. This effect may be specific to child-directed speech, which may be more grounded in the here and now than general adult speech, such that more frequently occurring lemmas are also more grounded. Additionally, concrete lemmas occur in more aligned utterances, suggesting that these words are more likely to occur in contexts in which the referred objects are also in the visual scene. 
The causal direction of these effects remains an open question---for example, it could be that utterances containing concrete words result in children's correct attending to the referred object, or it could be that having concrete objects in the environment drives adult to talk about those objects. In-lab interventional experiments (e.g., varying the number and concreteness of objects in the environment) could help to determine the direction of causality.

We anticipated that alignment would, on average, decline with age as speech becomes more abstract (``Are you feeling happy?'') and adults are more likely to refer to absent referents (e.g., ``Can you go get froggy?''). However, we did not find strong evidence for this trend in our dataset, as only our exploratory analyses revealed this effect for the longest utterances. However, a key avenue for future work is quantifying additional sources of variability in at-home egocentric recordings (e.g., adult- vs. child-directed speech, episodes of joint attention with caregivers) that may influence alignment. 

We suspect that assessing vision--language alignment within more flexible time windows (e.g., 2 seconds before or after an utterance) may lead to the recovery of some highly aligned moments that were not captured by the present analyses. More flexible time windows may help to identify the episodes that are likely to be most valuable for learning and that may be underestimated in our analyses. Analyzing longer time windows would also help to determine the role of alignment dynamics over the course of back-and-forth exchange, such as whether the caregiver or the child initiates a bid for joint attention \citep[see][]{bianco2025contingent}. Averaging across the embeddings of neighboring utterances and frames during such exchanges could lead to increased vision--language alignment as well \citep{he2025seeing}. 

One additional direction for future work relates to the fact that CLIP conducts a center crop on the image before encoding it; as such, information that lies outside of the center crop (including, notably, text on a book during shared book reading that may appear at the bottom of the visual field) would not be used to determine the CLIP alignment score. An initial exploration of alternative image processing steps, such as padding and squashing, to include information outside of the center crop did produce moderate differences in 4AFC accuracy. In future work, analyzing CLIP alignment scores between an utterance and different segments of the visual field will also allow us to understand the distribution of aligned visual information across the visual field.

More broadly, this work focused on at-home recordings of monolingual English-speaking families in the United States. We suspect that visual--linguistic alignment properties may differ depending on context---for example, due to cultural differences in how caregivers draw infants' attention to objects \citep{senzaki2016communication}. Collection and analyses of data from broader and more diverse populations would help to determine the generalizability of these findings.

Overall, our results suggest that CLIP alignment score is a meaningful metric to estimate the vision--language alignment in naturalistic egocentric videos, but that the alignment in young children's everyday environments is relatively infrequent compared to alignment in modern machine learning datasets. 
Nonetheless, this alignment varies in systematic ways based on speaker, utterance properties, and features of the lemmas in those utterances. Future work exploring the dynamics, distribution, and predictors of variation in alignment will help to better capture the complex landscape of children's early visual and linguistic input, and therefore inform our theories and models of language learning in young children.

\section{Acknowledgments}
We gratefully acknowledge the families who participated in the BabyView Dataset. This work was funded by an NIH R00HD108386 grant to B.L., by a grant from Schmidt Futures, by a gift from Meta, by the Stanford Center for the Study of Language and Information John Crosby Olney Fund, and by the Stanford Human-Centered AI Initiative (HAI) Hoffman-Yee grant program.

\bibliographystyle{apacite}

\setlength{\bibleftmargin}{.125in}
\setlength{\bibindent}{-\bibleftmargin}

\bibliography{ccn_style}

@article{vong2025robustness,
  title={On the robustness of modeling grounded word learning through a child's egocentric input},
  author={Vong, Wai Keen and Lake, Brenden M},
  journal={arXiv preprint arXiv:2507.14749},
  year={2025}
}

@article{long2024babyview,
  title={The BabyView dataset: High-resolution egocentric videos of infants' and young children's everyday experiences},
  author={Long, Bria and Sparks, Robert Z and Xiang, Violet and Stojanov, Stefan and Yin, Zi and Keene, Grace E and Tan, Alvin WM and Feng, Steven Y and Zhuang, Chengxu and Marchman, Virginia A and others},
  journal={arXiv preprint arXiv:2406.10447},
  year={2024}
}

@inproceedings{radford2021learning,
  title={Learning transferable visual models from natural language supervision},
  author={Radford, Alec and Kim, Jong Wook and Hallacy, Chris and Ramesh, Aditya and Goh, Gabriel and Agarwal, Sandhini and Sastry, Girish and Askell, Amanda and Mishkin, Pamela and Clark, Jack and others},
  booktitle={International conference on machine learning},
  pages={8748--8763},
  year={2021},
  organization={PMLR}
}

@article{yoshida2008s,
  title={What's in view for toddlers? Using a head camera to study visual experience},
  author={Yoshida, Hanako and Smith, Linda B},
  journal={Infancy},
  volume={13},
  number={3},
  pages={229--248},
  year={2008},
  publisher={Wiley Online Library}
}

@article{bergelson2017nature,
  title={Nature and origins of the lexicon in 6-mo-olds},
  author={Bergelson, Elika and Aslin, Richard N},
  journal={Proceedings of the National Academy of Sciences},
  volume={114},
  number={49},
  pages={12916--12921},
  year={2017},
  publisher={National Acad Sciences}
}

@article{vong2024grounded,
  title={Grounded language acquisition through the eyes and ears of a single child},
  author={Vong, Wai Keen and Wang, Wentao and Orhan, A Emin and Lake, Brenden M},
  journal={Science},
  volume={383},
  number={6682},
  pages={504--511},
  year={2024},
  publisher={American Association for the Advancement of Science}
}

@article{zhang2021cross,
  title={Cross-situational learning from ambiguous egocentric input is a continuous process: Evidence using the human simulation paradigm},
  author={Zhang, Yayun and Yurovsky, Daniel and Yu, Chen},
  journal={Cognitive science},
  volume={45},
  number={7},
  pages={e13010},
  year={2021},
  publisher={Wiley Online Library}
}

@article{yu2007rapid,
  title={Rapid word learning under uncertainty via cross-situational statistics},
  author={Yu, Chen and Smith, Linda B},
  journal={Psychological science},
  volume={18},
  number={5},
  pages={414--420},
  year={2007},
  publisher={SAGE Publications Sage CA: Los Angeles, CA}
}

@article{sullivan2021saycam,
  title = {{{SAYCam}}: {{A Large}}, {{Longitudinal Audiovisual Dataset Recorded From}} the {{Infant}}'s {{Perspective}}},
  shorttitle = {{{SAYCam}}},
  author = {Sullivan, Jessica and Mei, Michelle and Perfors, Andrew and Wojcik, Erica and Frank, Michael C.},
  year = 2021,
  month = may,
  journal = {Open Mind: Discoveries in Cognitive Science},
  volume = {5},
  pages = {20--29},
  issn = {2470-2986},
  doi = {10.1162/opmi_a_00039},
  urldate = {2022-11-27},
  pmcid = {PMC8412186},
  pmid = {34485795}
}

@inproceedings{lavechin2020open,
  title = {An {{Open-Source Voice Type Classifier}} for {{Child-Centered Daylong Recordings}}},
  booktitle = {Proceedings of {{Interspeech}} 2020},
  author = {Lavechin, Marvin and Bousbib, Ruben and Bredin, Herv{\'e} and Dupoux, Emmanuel and Cristia, Alejandrina},
  year = 2020,
  month = oct,
  pages = {3072--3076},
  address = {Shanghai, China},
  doi = {10.21437/Interspeech.2020-1690}
}

@article{sepuri2025characterizing,
  author = {Sepuri, T. and Aw, K. L. and Tan, A. W. M. and Sparks, R. Z. and Marchman, V. A. and Frank, M. C. and Long, B.},
  title = "Characterizing young children’s everyday activities using video question-answering models",
  year         = {2025},
  month        = {October 10},
journal={PsyArXiv preprint},
  doi          = {10.31234/osf.io/gndy9_v1},
  url          = {https://doi.org/10.31234/osf.io/gndy9_v1}
}

@inproceedings{straka2016udpipe,
    title = "{UDP}ipe: Trainable Pipeline for Processing {C}o{NLL}-{U} Files Performing Tokenization, Morphological Analysis, {POS} Tagging and Parsing",
    author = "Straka, Milan  and
      Haji{\v{c}}, Jan  and
      Strakov{\'a}, Jana",
    editor = "Calzolari, Nicoletta  and
      Choukri, Khalid  and
      Declerck, Thierry  and
      Goggi, Sara  and
      Grobelnik, Marko  and
      Maegaard, Bente  and
      Mariani, Joseph  and
      Mazo, Helene  and
      Moreno, Asuncion  and
      Odijk, Jan  and
      Piperidis, Stelios",
    booktitle = "Proceedings of the Tenth International Conference on Language Resources and Evaluation ({LREC}'16)",
    month = may,
    year = "2016",
    address = "Portoro{\v{z}}, Slovenia",
    publisher = "European Language Resources Association (ELRA)",
    url = "https://aclanthology.org/L16-1680/",
    pages = "4290--4297",
}

@article{brysbaert2014concreteness,
  title = {Concreteness ratings for 40 thousand generally known English word lemmas},
  volume = {46},
  ISSN = {1554-3528},
  url = {http://dx.doi.org/10.3758/s13428-013-0403-5},
  DOI = {10.3758/s13428-013-0403-5},
  number = {3},
  journal = {Behavior Research Methods},
  publisher = {Springer Science and Business Media LLC},
  author = {Brysbaert,  Marc and Warriner,  Amy Beth and Kuperman,  Victor},
  year = {2014},
  month = oct,
  pages = {904–911}
}

@article{cortese2004imageability,
  title = {Imageability ratings for 3,000 monosyllabic words},
  volume = {36},
  ISSN = {1532-5970},
  url = {http://dx.doi.org/10.3758/BF03195585},
  DOI = {10.3758/bf03195585},
  number = {3},
  journal = {Behavior Research Methods, Instruments, \& Computers},
  publisher = {Springer Science and Business Media LLC},
  author = {Cortese,  Michael J. and Fugett,  April},
  year = {2004},
  month = aug,
  pages = {384–387}
}

@article{schock2011imageability,
  title = {Imageability estimates for 3,000 disyllabic words},
  volume = {44},
  ISSN = {1554-3528},
  url = {http://dx.doi.org/10.3758/s13428-011-0162-0},
  DOI = {10.3758/s13428-011-0162-0},
  number = {2},
  journal = {Behavior Research Methods},
  publisher = {Springer Science and Business Media LLC},
  author = {Schock,  Jocelyn and Cortese,  Michael J. and Khanna,  Maya M.},
  year = {2011},
  month = oct,
  pages = {374–379}
}

@article{lynott2020lancaster,
  title = {The Lancaster Sensorimotor Norms: multidimensional measures of perceptual and action strength for 40, 000 English words},
  volume = {52},
  ISSN = {1554-3528},
  url = {http://dx.doi.org/10.3758/s13428-019-01316-z},
  DOI = {10.3758/s13428-019-01316-z},
  number = {3},
  journal = {Behavior Research Methods},
  publisher = {Springer Science and Business Media LLC},
  author = {Lynott,  Dermot and Connell,  Louise and Brysbaert,  Marc and Brand,  James and Carney,  James},
  year = {2020},
  month = dec,
  pages = {1271–1291}
}

@article{khanna2021how,
  title = {How well imageability,  concreteness,  perceptual strength,  and action strength predict recognition memory,  lexical decision,  and reading aloud performance},
  volume = {29},
  ISSN = {1464-0686},
  url = {http://dx.doi.org/10.1080/09658211.2021.1924789},
  DOI = {10.1080/09658211.2021.1924789},
  number = {5},
  journal = {Memory},
  publisher = {Informa UK Limited},
  author = {Khanna,  Maya M. and Cortese,  Michael J.},
  year = {2021},
  month = may,
  pages = {622–636}
}

@inproceedings{bianco2025contingent,
  title = {Contingent Behavior during Caregiver-Child Interaction Improves the Quality of Word Learning Opportunities},
  booktitle = {2025 {{IEEE International Conference}} on {{Development}} and {{Learning}} ({{ICDL}})},
  author = {Bianco, Catherine and Pang, Jingwen and Yu, Chen},
  year = 2025,
  month = sep,
  pages = {1--6},
  doi = {10.1109/ICDL63968.2025.11204427},
  urldate = {2025-10-30}
}

@misc{lin2015microsoft,
      title={Microsoft COCO: Common Objects in Context}, 
      author={Tsung-Yi Lin and Michael Maire and Serge Belongie and Lubomir Bourdev and Ross Girshick and James Hays and Pietro Perona and Deva Ramanan and C. Lawrence Zitnick and Piotr Dollár},
      year={2015},
      eprint={1405.0312},
      archivePrefix={arXiv},
      primaryClass={cs.CV},
      url={https://arxiv.org/abs/1405.0312}, 
}

@article{young2014image,
    title={From image descriptions to visual denotations: New similarity metrics for semantic inference over event descriptions},
    author={Peter Young and Alice Lai and Micah Hodosh and Julia Hockenmaier},
    journal={TACL},
    volume={2},
    pages={67--78},
    year={2014}
}

@inproceedings{he2025seeing,
  title={Seeing Through Words, Speaking Through Pixels: Deep Representational Alignment Between Vision and Language Models},
  author={He, Zoe Wanying and Trott, Sean and Khosla, Meenakshi},
  booktitle={Proceedings of the 2025 Conference on Empirical Methods in Natural Language Processing},
  pages={35645--35660},
  year={2025}
}

@article{senzaki2016communication,
  title = {The Communication of Culturally Dominant Modes of Attention from Parents to Children: A Comparison of Canadian and Japanese Parent-Child Conversations during a Joint Scene Description Task},
  volume = {11},
  ISSN = {1932-6203},
  url = {http://dx.doi.org/10.1371/journal.pone.0147199},
  DOI = {10.1371/journal.pone.0147199},
  number = {1},
  journal = {PLOS ONE},
  publisher = {Public Library of Science (PLoS)},
  author = {Senzaki,  Sawa and Masuda,  Takahiko and Takada,  Akira and Okada,  Hiroyuki},
  editor = {Kendal,  Rachel L},
  year = {2016},
  month = jan,
  pages = {e0147199}
}

@misc{radford2022robust,
      title={Robust Speech Recognition via Large-Scale Weak Supervision}, 
      author={Alec Radford and Jong Wook Kim and Tao Xu and Greg Brockman and Christine McLeavey and Ilya Sutskever},
      year={2022},
      eprint={2212.04356},
      archivePrefix={arXiv},
      primaryClass={eess.AS},
      url={https://arxiv.org/abs/2212.04356}, 
}

@misc{gandhi2023distil,
      title={Distil-Whisper: Robust Knowledge Distillation via Large-Scale Pseudo Labelling}, 
      author={Sanchit Gandhi and Patrick von Platen and Alexander M. Rush},
      year={2023},
      eprint={2311.00430},
      archivePrefix={arXiv},
      primaryClass={cs.CL},
      url={https://arxiv.org/abs/2311.00430}, 
}

@misc{jina2023clip,
    title={CLIP-as-service},
    author={{Jina AI}},
    year={2023},
    url={https://clip-as-service.jina.ai/}
}

@book{frank2021,
  title = {Variability and Consistency in Early Language Learning: The Wordbank Project},
  ISBN = {9780262362979},
  url = {http://dx.doi.org/10.7551/mitpress/11577.001.0001},
  DOI = {10.7551/mitpress/11577.001.0001},
  publisher = {The MIT Press},
  author = {Frank,  Michael C. and Braginsky,  Mika and Yurovsky,  Daniel and Marchman,  Virginia A.},
  year = {2021},
  month = mar 
}

@article{marchman2008,
  title = {Speed of word recognition and vocabulary knowledge in infancy predict cognitive and language outcomes in later childhood},
  volume = {11},
  ISSN = {1467-7687},
  url = {http://dx.doi.org/10.1111/j.1467-7687.2008.00671.x},
  DOI = {10.1111/j.1467-7687.2008.00671.x},
  number = {3},
  journal = {Developmental Science},
  publisher = {Wiley},
  author = {Marchman,  Virginia A. and Fernald,  Anne},
  year = {2008},
  month = may 
}

@article{frank2013,
  title = {Social and Discourse Contributions to the Determination of Reference in Cross-Situational Word Learning},
  volume = {9},
  ISSN = {1547-3341},
  url = {http://dx.doi.org/10.1080/15475441.2012.707101},
  DOI = {10.1080/15475441.2012.707101},
  number = {1},
  journal = {Language Learning and Development},
  publisher = {Informa UK Limited},
  author = {Frank,  Michael C. and Tenenbaum,  Joshua B. and Fernald,  Anne},
  year = {2013},
  month = jan,
  pages = {1–24}
}

@article{zhang2025videollama,
  title={{VideoLLaMA} 3: Frontier Multimodal Foundation Models for Image and Video Understanding},
  author={Zhang, Boqiang and Li, Kehan and Cheng, Zesen and Hu, Zhiqiang and Yuan, Yuqian and Chen, Guanzheng and Leng, Sicong and Jiang, Yuming and Zhang, Hang and Li, Xin and Jin, Peng and Zhang, Wenqi and Wang, Fan and Bing, Lidong and Zhao, Deli},
  journal={arXiv preprint arXiv:2501.13106},
  year={2025},
  url = {https://arxiv.org/abs/2501.13106}
}

\end{document}